\pdfoutput=1

\documentclass[11pt]{article}

\usepackage{acl}
\usepackage{amsmath}
\usepackage{adjustbox}
\usepackage{graphicx}
\usepackage{booktabs}
\usepackage{multirow}
\usepackage{multicol}
\usepackage{makecell}
\usepackage{enumitem}
\usepackage{times}
\usepackage{latexsym}
\usepackage{multirow}

\usepackage[T1]{fontenc}

\usepackage[utf8]{inputenc}

\usepackage{microtype}

\usepackage{inconsolata}

\usepackage{graphicx}

\usepackage[symbol]{footmisc}

\title{Retrieval, Reasoning, Re-ranking: A Context-Enriched Framework for Knowledge Graph Completion}

\author{
 \textbf{Muzhi Li\textsuperscript{2,1}}\footnotemark[1],
 \textbf{ Cehao Yang\textsuperscript{3,1}}\footnotemark[1],
 \textbf{ Chengjin Xu\textsuperscript{1}}\footnotemark[1]\footnotemark[2],
 \textbf{ Xuhui Jiang\textsuperscript{1}},\\
 \textbf{Yiyan Qi\textsuperscript{1}},
 \textbf{ Jian Guo\textsuperscript{1}}\footnotemark[2],
 \textbf{ Ho-fung Leung}\footnotemark[3],
 \textbf{ Irwin King\textsuperscript{2}}\footnotemark[2]
\\
 \textsuperscript{1}IDEA Research, International Digital Economy Academy
 \\
 \textsuperscript{2}Department of Computer Science \& Engineering, The Chinese University of Hong Kong
 \\
 \textsuperscript{3}Artificial Intelligence Thrust, Hong Kong University of Science and Technology (Guangzhou)
\\
\texttt{\{limuzhi,yangcehao,xuchengjin,jiangxuhui,qiyiyan,guojian\}@idea.edu.cn}
\\
\texttt{\{mzli,king\}@cse.cuhk.edu.hk}, \texttt{ho-fung.leung@outlook.com} \\
\texttt{cyang289@connect.hkust-gz.edu.cn}
}

\DeclareMathOperator*{\argmax}{arg\,max}

\begin{document}
\maketitle
\footnotetext[1]{Equal contribution.}
\footnotetext[2]{Corresponding authors.}
\footnotetext[3]{Independent researcher.}
\begin{abstract}
The Knowledge Graph Completion~(KGC) task aims to infer the missing entity from an incomplete triple. Existing embedding-based methods rely solely on triples in the KG, which is vulnerable to specious relation patterns and long-tail entities. On the other hand, text-based methods struggle with the semantic gap between KG triples and natural language. Apart from triples, entity contexts (e.g., labels, descriptions, aliases) also play a significant role in augmenting KGs. To address these limitations, we propose $\text{KGR}^3$, a context-enriched framework for KGC. $\text{KGR}^3$ is composed of three modules. Firstly, the \textit{Retrieval} module gathers supporting triples from the KG, collects plausible candidate answers from a base embedding model, and retrieves context for each related entity. Then, the \textit{Reasoning} module employs a large language model to generate potential answers for each query triple. Finally, the \textit{Re-ranking} module combines candidate answers from the two modules mentioned above, and fine-tunes an LLM to provide the best answer. Extensive experiments on widely used datasets demonstrate that $\text{KGR}^3$ consistently improves various KGC methods. Specifically, the best variant of $\text{KGR}^3$ achieves absolute Hits@$1$ improvements of $12.3\%$ and $5.6\%$ on the FB15k237 and WN18RR datasets.
\end{abstract}

\section{Introduction}
Knowledge Graphs~(KGs) are graph-structured knowledge bases~(KBs) that organize factual knowledge as triples in the form of (\textit{head entity}, \textit{relation}, \textit{tail entity}).
Recently, KGs have become a crucial foundation for various downstream applications, such as recommendation systems~\cite{lkgr,chen2022attentive}, question answering~\cite{ToG}, and sentiment analysis~\cite{wang-shu-2023-explainable}. Nevertheless, mainstream KGs such as Freebase~\cite{Freebase} and Wordnet~\cite{WordNet} suffer from serious incomplete issues. This problem highlights the importance of the \textbf{\textit{Knowledge Graph Completion}}~(KGC) task, which aims to predict the missing entity from an incomplete triple.

\begin{figure}
    \vspace{0.15cm}
    \centering
    \includegraphics[width=0.46\textwidth]{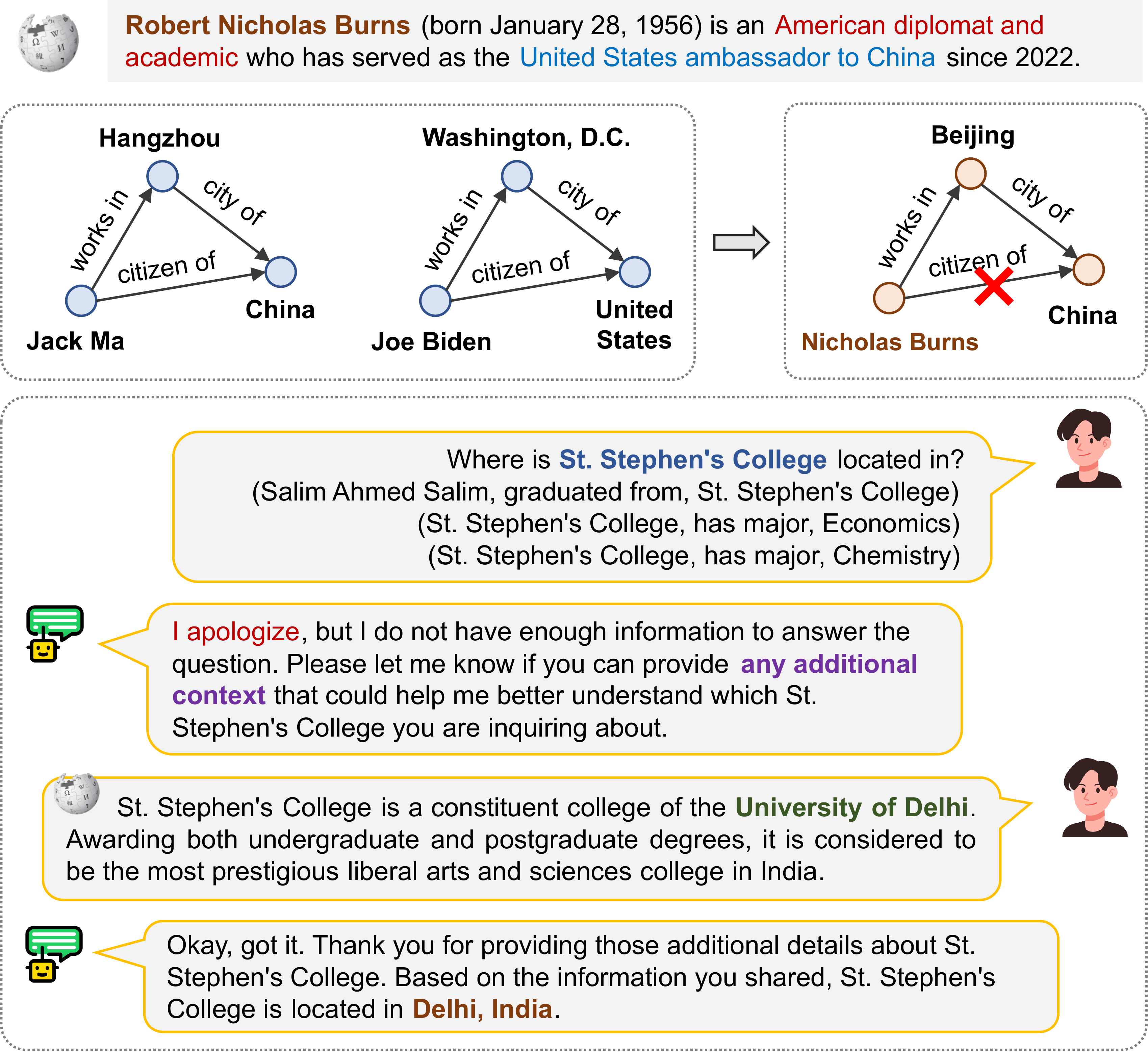}
    \caption{Limitations of existing embedding-based (top) and LLM-based (bottom) KGC methods. }
    \label{fig:kg}
\end{figure}

Existing KGC methods can be roughly categorized into embedding-based methods~\cite{TransE,DistMult,RotatE,GIE} and text-based methods~\cite{kgbert,KEPLER,SimKGC}. Embedding-based methods 
implicitly learn rules based on relation patterns observed in triples and make predictions based on the likelihood of these patterns occurring~\cite{Rule}. For example, from triple $(X, \textit{works in}, Y)$ and $(Y, \textit{city of}, Z)$, it is very likely to deduce that $(X, \textit{citizen of}, Z)$. However, these methods ignore the contextual semantics supporting these triples, leading to conclusions that do not align with the facts. Text-based methods employ pre-trained language models~(PLMs) to embed entities and relations with their labels and descriptions. However, these methods still cannot surpass the latest embedding-based counterparts~\cite{CompoundE} due to \textbf{\textit{the substantial semantic gap between structural KG triples and natural language sentences}}.

Large language models~(LLMs), trained by extensive corpora, demonstrate emergent semantic understanding and in-context learning~(ICL) capabilities. Recent studies \cite{KICGPT,DIFT} have proposed utilizing LLMs for the KGC task, as these models harbor general knowledge that can be leveraged to mitigate information scarcity for long-tail entities. However, the application of LLMs in KGC tasks encounters several limitations. Firstly, if the pre-training corpora of the LLMs lack adequate contextual information on specific entities, the LLMs may produce hallucinated or biased responses. Secondly, the structured nature of KG triples limits the ability of LLMs to effectively capture and leverage contextual information from the graph structure. These shortcomings necessitate a strong reliance on a considerable amount of in-context demonstrations~\cite{KICGPT} or external structured embeddings~\cite{DIFT}, which inevitably limit the performance and generality of existing approaches. 

Considering the aforementioned challenges, we propose a context-enriched KGC framework named $\textbf{KGR}^{\mathbf{3}}$, which consists of three modules: \textit{\textbf{R}etrieval}, \textit{\textbf{R}easoning}, and \textit{\textbf{R}e-ranking}. Given a query triple, the \textit{retrieval} module gathers semantically relevant supporting triples with the same relation and similar entities, and extracts plausible candidate answers from a base KGC model. To ensure that the LLM attains a fundamental understanding of the retrieved information, this module also collects and augments relevant contextual information to entities in supporting triples and the candidate answer list. Then, the \textit{reasoning} module exploits the semantic understanding capability of pre-trained LLM to suggest several potential answers based on in-context demonstrations and the description of the known entity. 
Finally, the re-ranking module fine-tunes the LLM to select out the corrupted entity of the training triple from a set of candidate entities, enabling it to process structured knowledge.
$\textbf{KGR}^\mathbf{3}$ possesses strong plug-and-play capability, making it compatible with all base KGC methods without costly re-training. During inference, the re-ranking module integrates the candidate answers derived from the base KGC model and the reasoning module, and then instructs the LLM to output the entity label that best completes the query triple. 

We validate the proposed framework on two conventional KGC datasets: FB15k237 and WN18RR. The extensive experiments show that $\text{KGR}^3$ significantly and consistently outperforms all baseline methods with different types of base KGC models and backbone LLMs, showing its superiority. Notably, the best variant of $\text{KGR}^3$ achieves state-of-the-art performance with absolute Hits@$1$ improvements of $12.3\%$ and $5.6\%$ on the two datasets. Our contributions are summarized as follows: 
\begin{itemize}[leftmargin=1em]
    \setlength\itemsep{0.1em}
    \item We propose a novel $\text{KGR}^3$ framework for the KGC task, which systematically retrieves relevant supporting contexts, conducts semantic reasoning, and re-ranks candidate answers.  

    \item We notice the semantic gap between KG triples and natural language sentences, and seamlessly bridge this gap with entity contexts. 
   
    \item We conduct extensive experiments and ablation studies to evaluate the effectiveness of the $\text{KGR}^3$ framework, and discuss the importance of incorporating entity contexts and LLMs. 
\end{itemize}

\section{Related Works}
\paragraph{Embedding-based methods}
Embedding-based methods are fundamental in KGC, which focuses on learning a set of low-dimensional embeddings for entities and relations with certain geometric or mathematical constraints. Most typically, TransE~\cite{TransE} assumes the translated head embedding of a triple is close to the embedding of the tail. DistMult~\cite{DistMult} aims to maximize the Hadamard product of the head, relation, and tail embeddings of each triple. To model symmetric and anti-symmetric relations, ComplEx~\cite{Complex} generalizes DistMult by introducing complex embeddings with Hermitian dot product. RotatE~\cite{RotatE} interprets relation as a rotation operation in complex space, which can effectively infer inversion and composition patterns. ATTH~\cite{ATTH} and GIE~\cite{GIE} further leverage hyperbolic embeddings and operations to capture the intrinsic hierarchical structure in KGs. 
Recently, Graph Neural Networks~(GNNs) have emerged as powerful methods for graph embedding~\cite{GNN,DBLP:conf/nips/SongZK23a,DBLP:conf/nips/SongZK23}, with various applications~\cite{DBLP:conf/aaai/MaSHLZK23,song2022individual,DBLP:conf/aaai/SongMK24} . Considering the heterogeneity of KGs, RGCN~\cite{RGCN}, WGCN~\cite{WGCN}, and CompGCN~\cite{CompGCN} adapted to consider relation types in their message-passing functions. KBGAT~\cite{KBGAT} proposes a two-layer attentional network to encode each triple and to measure its importance to the tail entity. 
In addition, NBF-Net~\cite{NBF-Net}, RED-GNN~\cite{RED-GNN} and A*Net~\cite{ANet} integrate structural information from paths between the two entities, which also supports inductive KGC. 
Despite the simplicity and high scalability, embedding methods suffer from the long-tail entity distribution in KGs, and hence, cannot generate semantic expressive embeddings for boundary entities, which limits the performance of KGC.

\paragraph{Text-based methods. }
Apart from the graph structure, textual information in KGs also entails rich semantic knowledge. 
DKRL proposes to initialize entity embeddings with a convolutional neural network. 
KG-BERT~\cite{kgbert} tokenizes triples with textual descriptions of entities and relations, and utilizes BERT~\cite{BERT} to assess their plausibility.   KEPLER~\cite{KEPLER} jointly finetunes the pre-trained BERT with the KG embedding and MLM objectives, showcasing improved KGC accuracy. SimKGC~\cite{SimKGC} introduces a contrastive learning strategy, which reduces the computational complexity by re-using in-batch and pre-batch entities as negative samples. CoLE~\cite{CoLE} extends KG-BERT with co-distillation learning. Inspired by the semantic understanding and reasoning capability of LLMs, KICGPT~\cite{KICGPT} proposes a GPT-based in-context learning~(ICL) paradigm for the KGC task. 
However, due to the large semantic gap between KG triples and natural language sentences, all approaches mentioned above cannot outperform the latest embedding-based methods~\cite{CompoundE,MGTCA}.
Recently, DIFT~\cite{DIFT} devises a supervised fine-tuning~(SFT) solution to guide the LLM in completing triples. Despite achieving state-of-the-art performance, DIFT necessitates costly re-training to adapt to different base KGC models, thereby limiting its compatibility and generality. 

\section{Problem Specifications}
A Knowledge Graph (denoted as $\mathcal{G} = \{\mathcal{E}, \mathcal{R}, \mathcal{T}\}$) can be represented as a set of triples in the form of $(h, r, t)\in \mathcal{T}$, where $h, t \in \mathcal{E}$, $r \in \mathcal{R}$. 
The notations $h$ and $t$ denote the head and the tail entity of a triple. $\mathcal{E}, \mathcal{R},\mathcal{T}$ are the set of entities, relations, and triples, respectively. Besides, KGs are usually associated with knowledge bases~(KBs), such as YAGO~\cite{YAGO}, Wikidata~\cite{Wikidata}, and DBPedia~\cite{DBPedia}. KBs are renowned for representing general knowledge about real-world objects, including people, organizations, places, products, and among others. Apart from triples, KBs also store rich contextual information (\textit{or ``contexts''}) for entities in KGs, which includes \textit{entity labels}, \textit{entity descriptions}, \textit{aliases}, etc. We argue that these entity contexts contain valuable semantic knowledge for the KGC task. In this paper, we exploit LLMs to reveal the missing entity of an incomplete triple based on relevant triples and the contexts of entities involved. 

\section{Methods}
In this section, we introduce our proposed context-enriched KGC framework~$\text{KGR}^3$, which consists of three components: (1) \textit{Retrieval}, (2) \textit{Reasoning}, and (3) \textit{Re-ranking}. 

\begin{figure*}
    \centering
    \includegraphics[width=\textwidth]{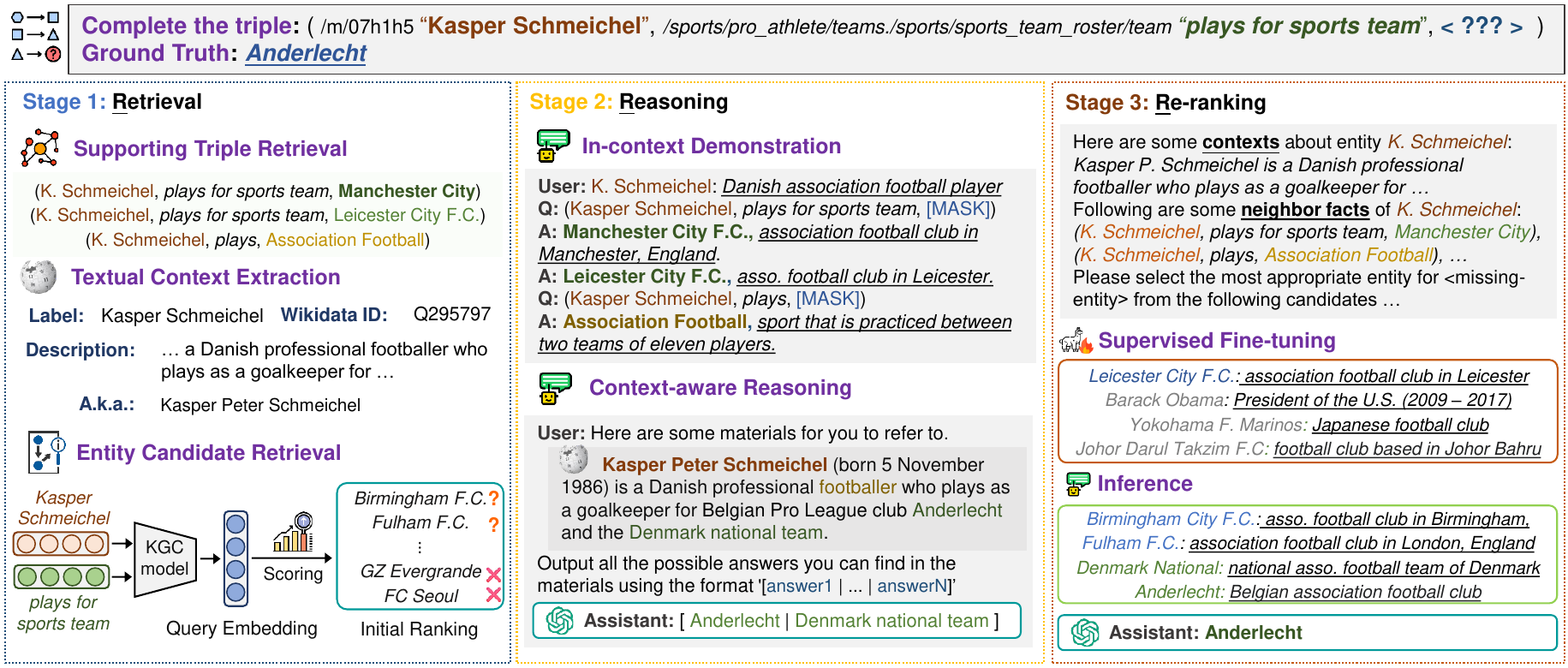}
    \caption{The end-to-end architecture of the proposed $\text{KGR}^3$ framework, which consists of three modules: \\1) Reasoning (left), 2) Reasoning (middle), and 3) Re-ranking (right). (Please see Appendix~\ref{prompts} for detailed prompts.)}
    \label{KGCFigure}
\end{figure*}

\subsection{Stage 1: Retrieval}
\label{retrieval}
The retrieval module focuses on gathering structural and semantic knowledge that may contribute to the completion of certain incomplete triples.
\subsubsection{Supporting Triple Retrieval }
In KGs, the attributes of an entity are represented in structural triples.  Different entities connected by the same relation often share common salient properties. The internal knowledge inherent in the graph structure provides the most direct support to the validity of a triple. Given an incomplete query triple in the form of $(h, r, ?)$ or $(?, r, t)$, we aim to retrieve $k$ supporting triples that are the most semantically similar to the incomplete query triple. Intuitively, we prioritize triples with the same entity and relation from the training set. If the number of available triples is less than $k$, we broaden our choices to triples with the same relation, and with entities that are semantically similar to the known one in the query triple. 

\subsubsection{Textual Context Retrieval}
We note that there is a significant semantic gap between structural triples and natural language sentences. For example, in Figure~\ref{KGCFigure}, entity ``Kasper Schmeichel'' is originally represented by an unique entity id ``\textit{/m/07h1h5}'' while relation ``plays for sports teams'' is originally represented as \textit{``/sports/pro\_athlete/teams\allowbreak ./sports/sports\_team\allowbreak \_roster/team''}. Such a structured format is difficult for LLMs to process. 
To fully leverage the semantic understanding capabilities of LLMs, we extract relevant contexts related to entities in the query triple and supporting triples from Wikidata knowledge base~\cite{Wikidata}. 

In mainstream KGs, entities are represented in numerical or textual IDs. Each entity ID acts as an index to the data frame in its corresponding knowledge base. Apart from triples, the data-frame of an entity contains significant contextual information such as its entity label. 
Since Google Freebase~\cite{Freebase} is deprecated and migrated to Wikidata~\cite{Wikidata}, we map the entity IDs in the FB15k237 dataset to corresponding Wikidata QIDs with official data dumps. We then collect the textual entity label, the short description, and aliases from Wikidata URIs. As for entities in the WN18RR dataset, we adopt the same set of entity labels and descriptions used in KGBERT~\cite{kgbert} and SimKGC~\cite{SimKGC}. 

\subsubsection{Candidate Answer Retrieval } 
The widely adopted ranking-based evaluation for the KGC task requires the model to score the plausibility of each entity in the KG as a potential replacement for the missing entity in the query triple. However, given the vast number of entities in the KG, employing LLMs to score and rank each entity is computationally expensive and impractical. 
Inspired by~\cite{RobustKGC,KICGPT,SSET}, we employ a base KGC model to initialize the scoring and ranking of entities within the KG. Formally, we denote the ranked entity list $\mathcal{A}_{\text{KGC}}$ as following: 
\begin{align}
    \mathcal{A}_{\text{KGC}} &= [e_1^{(k)}, e_2^{(k)}, ..., e_{|\mathcal{E}|}^{(k)}], \\
    \text{ where } e_i^{(k)} &= \argmax_{e\in \{\mathcal{E}\backslash \{e_{<i}^{(k)}\}\}}{f_r(h, e) \text{ or } f_r(e, t)}, \notag
\end{align}
$f_r(h, t)$ is the scoring function the KGC model evaluating the plausibility of a triple $(h,r,t)$. 
Then, we extract the top-$n$ entities with the highest scorings from $\mathcal{A}_{\text{KGC}}$ as candidate answers, and extract their labels and descriptions based on their entity IDs. The plausibility of these candidate answers will be re-evaluated in the re-ranking module.

\subsection{Stage 2: Reasoning}
\label{reasoning}
In the second stage, we first exploit the supporting triples to direct the LLM in performing the KGC task. In addition, we employ the LLM to generate several possible answers drawing upon the description of the known entity in the query triple.  
\subsubsection{Supporting Triple Demonstrations }
In this sub-section, we present the design of our prompt tailored for the demonstration, which is crucial in facilitating in-context learning~\cite{KICGPT}.  For each supporting triple, we first provide LLM with the description of the head entity. The entity description serves three objectives: (1)~\textit{disambiguate entities with the same label}, (2) \textit{rejuvenate the LLM’s memory about known entities}, and (3) \textit{provide essential information for entities that are not included in the LLM pre-training corpus}. Following the convention of~\cite{AutoKG,KICGPT}, we corrupt the neighboring entity with a ``[MASK]'' token. In order to narrow the semantic gap between structural triples and natural language sentences, we translate the masked triple into a natural language question. Subsequently, we ask the LLM to generate answers based on its semantic understanding of the short description of the known entity and the question. Different from KICGPT~\cite{KICGPT}, all supporting triples included in our demonstration prompts are sampled from the training set, which avoids potential information leakage. 
Finally, we provide the label of the corrupted entity and its entity description as our explanation. Considering multiple supporting triples, the LLM can also discern salient properties of adjacent neighbors connected by the same relation and similar entities. These salient properties play a vital role in helping the LLM to find out proper answers for the query triple. 

\subsubsection{Context-aware Reasoning }
We anticipate that LLMs can harness their information extraction and semantic understanding capabilities by utilizing comprehensive contextual information about the known entity, thereby generating potential answers. Similarly, we pass the description of the known entity
and the question translated from the query triple to the LLM. The LLM is then instructed to output a list of answers in its response. It should be noted that generative LLMs do not guarantee that output answers will conform to entities in the KG. Therefore, we post-process the LLM output by replacing entity aliases with entity labels and filtering out invalid and unreliable answers that do not appear within the top-$\delta$ positions of $\mathcal{A}_{\text{KGC}}$. Finally, we obtain a list of $m$ answers, which are formally denoted as: 
\begin{align}
    \mathcal{A}_{\text{LLM}} &= [e_1^{(l)}, e_2^{(l)}, ..., e_m^{(l)}] \notag \\
    &= f_{\text{LLM}}(q, c_e^{(q)}, \mathcal{D}(q)) \cap \mathcal{A}_{\text{KGC}}[0:\delta],
\end{align}
where $q, c_e^{(q)}$ and $\mathcal{D}(q)$ denote the question, description of the known entity, and supporting triple demonstrations. Entities in $\mathcal{A}_{\text{LLM}}$ are ensured to be simultaneously supported by the LLM and the base KGC model. 

\subsection{Stage 3: Re-ranking}
\label{reranking}
Motivated by the complementary nature of semantic and structural knowledge, we aim to exploit the candidate answer list generated by the LLM and the base KGC model to compose our final rankings. 

To better enable the LLM in utilizing entity descriptions and structured neighbor facts for ranking candidate answers to query triples, we introduce supervised fine-tuning~(SFT) with LoRA adaptation~\cite{LLMRanker}. Inspired by DIFT~\cite{DIFT}, the training objective of SFT is to find out the missing entity of an incomplete triple from a set of candidate answers. Specifically, we construct training samples by corrupting the tail (or head) entity of each triple in the training set. For each corrupted triple, we randomly sample $n-1$ negative samples from the entity set, where half of them are connected by the same relation as the corrupted ground truth entity. Incorporating these hard negative samples helps the LLM to distinguish between different entities with the same property, which is crucially important since candidate entities suggested by base KGC models usually yield similar characteristics. After that, we translate the masked triple to a natural language question, and gather the label and description for each candidate entity. 
Finally, we provide the question $q$, retrieved neighbor facts $\mathcal{N}(q)$, and the description of the known entity $c_e^{(q)}$, along with candidate answers $\mathcal{A}$ and their descriptions $c(\mathcal{A})$ to the LLM, and fine-tune the LLM to output the label $y$ of the ground truth entity. Formally, we have the SFT loss:
\begin{equation}
    \mathcal{L}_{\text{SFT}} = -\sum_{i=1}^{|\mathcal{T}|}\log(y|q, \mathcal{N}(q), c_e^{(q)},\mathcal{A}, c(\mathcal{A})).
\end{equation}
It is important to emphasize that the SFT process does not rely on the prior inference results from existing KGC approaches. This ensures that the $\text{KGR}^3$ framework can be implemented as a plug-and-play solution.   

During the inference stage, we construct a candidate answer set $\mathcal{A}_c$, which is composed of top-$p$ entities from $\mathcal{A}_{\text{KGC}}$ and top-$(n-p)$ entities in $\mathcal{A}_{\text{LLM}}$ that are not previously encountered. Formally, we have:
\begin{equation}
\resizebox{.99\linewidth}{!}{$
    \mathcal{A}_c = \mathcal{A}_{\text{KGC}}[0:p] \cup \{\mathcal{A}_{\text{LLM}}\backslash \mathcal{A}_{\text{KGC}}[0:p]\}[0:n-p]. $}
\end{equation}
If $\mathcal{A}_{\text{LLM}}$ contains fewer than $(n-p)$ entities (e.g. $m < n-p$), we supplement the candidate answer set with additional entities from $\mathcal{A}_{\text{KGC}}$ to reach a total of $n$ entities.
Similarly, we instruct the fine-tuned LLM to select the most appropriate candidate entity to complete the query triple, and output its entity label. The selected entity $e_{\text{ans}}$ is then prioritized and moved to the front of the candidate answer list $\mathcal{A}_c$. Finally, we construct the re-ordered entity list $\mathcal{A}_{\text{RR}}$ for performance evaluation. Here, we have
\begin{equation}
    \mathcal{A}_{\text{RR}} = \left[e_{\text{ans}} || \mathcal{A}_c\backslash\{e_{\text{ans}}\} || \mathcal{A}_{\text{KGC}}[n:|\mathcal{E}|] \right],
\end{equation}
where $[\cdot||\cdot]$ denotes the concatenation operation. 

\section{Experiments}
In this section, we assess the effectiveness of the $\text{KGR}^3$ framework in the KGC task. 

\subsection{Datasets}
We utilize two widely-used benchmark datasets, namely FB15k237 and WN18RR to evaluate the proposed method. FB15k237 is derived from Freebase~\cite{Freebase}, an encyclopedic KG containing general knowledge about topics such as celebrities, organizations, movies and sports. WN18RR is a subset of WordNet~\cite{WordNet}, a lexical KG with knowledge about English morphology. To prevent potential data leakage, FB15k237 and WN18RR excludes reversible relations from their backend KGs. Detailed statistics of the two datasets are shown in Table~\ref{dataset}. 
\begin{table}[htbp]\centering
\small
\begin{tabular}{ccc}
\toprule
\textbf{Dataset}& \textbf{FB15k237} & \textbf{WN18RR}\\
\midrule
\#Entities & 14,541 & 40,943 \\
\#Relations & 237 & 11 \\
\#Train & 272,115 & 86,835 \\
\#Valid & 17,535 & 3,034 \\
\#Test & 20,466 & 3,134 \\
\bottomrule
\end{tabular}
\caption{Statistics of Datasets}
\label{dataset}
\vspace{-0.15cm}
\end{table}

\subsection{Baselines and Evaluation Metrics} We compare the $\text{KGR}^3$ framework with four types of baseline methods: (1) traditional KG Embedding methods TransE~\cite{TransE}, ComplEx~\cite{Complex}, RotatE~\cite{RotatE}, TuckER~\cite{TuckER}; (2) GNN-based embedding methods CompGCN~\cite{CompGCN}, NBF-Net~\cite{NBF-Net}; (3) text-based methods KG-BERT~\cite{kgbert}, MEM-KGC~\cite{MEM-KGC}, SimKGC~\cite{SimKGC}, CoLE~\cite{CoLE}; and (4) LLM-based methods KICGPT~\cite{KICGPT}, DIFT~\cite{DIFT}. Among these baselines, we select TransE, RotatE, GIE, SimKGC, CoLE, and NBF-Net as our base KGC models because these methods are highly representative and exhibits strong performance. 

We utilize the widely adopted evaluation metrics, namely Hits@$k$~($k=1,3,10$) and MRR to evaluate our proposed method. Hits@$k$ measures the proportion of query triples which the ground truth entities are ranked within the top-$k$ position. MRR measures the mean reciprocal rank for each ground truth entities. Higher results indicates a better performance. Implementation details of baseline models and our models are described in Appendix~\ref{sec:implementation}.

\begin{table*}[t]
\centering
\small
\begin{tabular}{l|cccc|cccc}
\toprule
\multicolumn{1}{l|}{\multirow{2}{*}{\textbf{Methods}}}      & \multicolumn{4}{c|}{\textbf{FB15k237}}       & \multicolumn{4}{c}{\textbf{WN18RR}}         \\
& \textbf{MRR}    & \textbf{Hits@1} & \textbf{Hits@3} & \textbf{Hits@10} & \textbf{MRR}    & \textbf{Hits@1} & \textbf{Hits@3} & \textbf{Hits@10} \\
\midrule
\multicolumn{9}{c}{\textit{KG embedding methods}} \\
TransE        & 0.279  & 0.198  & 0.376  & 0.441   & 0.243  & 0.043  & 0.441  & 0.532   \\
ComplEx       & 0.247  & 0.158  & 0.275  & 0.428   & 0.440   & 0.410   & 0.460   & 0.510    \\
RotatE        & 0.338  & 0.241  & 0.375  & 0.533   & 0.476  & 0.428  & 0.492  & 0.571   \\
TuckER        & 0.358  & 0.266  & 0.394  & 0.544   & 0.470   & 0.443  & 0.482  & 0.526   \\
GIE           & 0.362  & 0.271  & 0.401  & 0.552   & 0.491  & 0.452  & 0.505  & 0.575   \\
HittER        & 0.373  & 0.279  & 0.409  & 0.558   & 0.503  & 0.462  & 0.516  & 0.584   \\
\midrule
\multicolumn{9}{c}{\textit{Graph neural network-based methods}} \\
CompGCN       & 0.355  & 0.264  & 0.390   & 0.535   & 0.479  & 0.443  & 0.494  & 0.546   \\
NBF-Net        & 0.415  & 0.321  & 0.450   & 0.599   & 0.551  & 0.497  & 0.573  & 0.666   \\
\midrule
\multicolumn{9}{c}{\textit{Text-based methods}} \\
KG-BERT       & -      & -      & -      & 0.420    & 0.216  & 0.041  & 0.302  & 0.524   \\
MEM-KGC       & 0.346  & 0.253  & 0.381  & 0.531   & 0.557  & 0.475  & 0.604  & 0.704   \\
SimKGC        & 0.338  & 0.252  & 0.364  & 0.511   & 0.671  & 0.595  & 0.719  & 0.802   \\
CoLE          & 0.389  & 0.294  & 0.429  & 0.572   & 0.593  & 0.538  & 0.616  & 0.701   \\
\midrule
\multicolumn{9}{c}{\textit{Large language model-based methods}} \\
ChatGPT (0-shot) & - & 0.237  & -  & -   & -  & 0.190  & -  & -   \\
ChatGPT (1-shot) & - & 0.267  & -  & -   & -  & 0.212  & -  & -   \\
KICGPT        & 0.412  & 0.327  & 0.448  & 0.581   & 0.564  & 0.478  & 0.612  & 0.677   \\
DIFT + TransE & 0.389  & 0.322   & 0.408  & 0.525   & 0.491  & 0.462  & 0.496  & 0.560    \\
DIFT + SimKGC & 0.402  & 0.338  & 0.418  & 0.528   & \underline{0.686}  & \underline{0.616}  & \underline{0.730}   & \underline{0.806}   \\
DIFT + CoLE   & 0.439  & 0.364  & 0.468  & 0.586   & 0.617  & 0.569  & 0.638  & 0.708   \\
\midrule
\textbf{$\textbf{KGR}^\mathbf{3}$ + TransE} & 0.456  & 0.414  & 0.474 & 0.550    & 0.506 & 0.487 & 0.515  & 0.556   \\
\textbf{$\textbf{KGR}^\mathbf{3}$ + RotatE} & 0.456  & 0.400 & 0.476  & 0.569   & 0.520 & 0.495 & 0.520   & 0.550    \\
\textbf{$\textbf{KGR}^\mathbf{3}$ + GIE}    & 0.463 & 0.400 & 0.485  & 0.581  & 0.558  & 0.520 & 0.580 & 0.615   \\
\textbf{$\textbf{KGR}^\mathbf{3}$ + SimKGC} & 0.471  & 0.429  & 0.490   & 0.557   & \textbf{0.717} & \textbf{0.656}   & \textbf{0.759}  & \textbf{0.809}   \\
\textbf{$\textbf{KGR}^\mathbf{3}$ + CoLE}   & \underline{0.507}  & \underline{0.455}  & \underline{0.537} & \underline{0.612}  & 0.635 & 0.579  & 0.676  & 0.723   \\
\textbf{$\textbf{KGR}^\mathbf{3}$ + NBF-Net} & \textbf{0.535} & \textbf{0.475} & \textbf{0.564} & \textbf{0.635}   & 0.641  & 0.605  & 0.662  & 0.695  \\
\bottomrule
\end{tabular}
\caption{Experiment results of the KGC task on FB15k-237 and WN18RR datasets. The best results are in \textbf{bold} and the second-best ones are \underline{underlined}. All results of baseline methods are referred from corresponding original papers. For $\text{KGR}^3$, we adopt LLama3-8B as the backbone LLM of the Reasoning and Re-ranking module. }
\label{tab:results}
\end{table*}

\begin{table*}[t]
\centering
\small
\begin{tabular}{l|cccc|cccc}
\toprule
\textbf{} & \multicolumn{4}{c|}{\textbf{FB15k237}} & \multicolumn{4}{c}{\textbf{WN18RR}} \\
\textbf{Settings} & \textbf{MRR} & \textbf{Hits@1} & \textbf{Hits@3} & \textbf{Hits@10} & \textbf{MRR} & \textbf{Hits@1} & \textbf{Hits@3} & \textbf{Hits@10} \\
\midrule
\textbf{$\textbf{KGR}^\mathbf{3}$ (Llama3-8B)} & \textbf{0.535} & \textbf{0.475} & \textbf{0.564} & \textbf{0.635} & \textbf{0.717} & \textbf{0.656} & \textbf{0.759} & \textbf{0.809} \\
w/o. Reasoning & 0.531 & 0.472 & 0.559 & 0.629 & 0.674 & 0.596 & 0.725 & 0.804 \\
w/o. Entity Descriptions & 0.523 & 0.460 & 0.554 & 0.627 & 0.675 & 0.601 & 0.723 & 0.801 \\
w/o. Neighbor Facts & 0.405 & 0.295 & 0.467 & 0.590 & 0.646 & 0.519 & 0.753 & 0.807 \\
\bottomrule
\end{tabular}
\caption{Results for ablation studies with the removal of reasoning module, entity descriptions, or neighbor facts. We adopt NBF-Net and SimKGC as the base KGC models for the FB15k237 and WN18RR datasets, respectively.  }
\label{tab:ablation_module}
\vspace{0.2cm}
\end{table*}

\begin{table*}[t]
\centering
\footnotesize
\begin{tabular}{ll|cccc|cccc}
\toprule
\multicolumn{2}{c|}{\textbf{LLMs}} & \multicolumn{4}{c|}{\textbf{FB15k237}} & \multicolumn{4}{c}{\textbf{WN18RR}} \\
\textbf{Re-ranking} & \textbf{Reasoning} & \textbf{MRR} & \textbf{Hits@1} & \textbf{Hits@3} & \textbf{Hits@10} & \textbf{MRR} & \textbf{Hits@1} & \textbf{Hits@3} & \textbf{Hits@10} \\
\midrule
Llama2-7B & Llama2-7B & 0.524 & 0.462 & 0.555 & 0.627 & 0.709 & 0.645 & 0.754 & 0.803 \\
Llama2-7B & GPT3.5 & 0.530 & 0.466 & 0.562 & 0.635 & 0.710 & 0.644 & 0.757 & \underline{0.808} \\
Llama3-8B & Llama3-8B & 0.535 & 0.475 & 0.564 & 0.634 & \underline{0.717} & \underline{0.656} & \textbf{0.759} & \textbf{0.809} \\
Llama3-8B & GPT3.5 & 0.536 & 0.477 & 0.565 & 0.636 & \underline{0.717} & 0.655 & \textbf{0.759} & 0.807 \\
\midrule
Qwen2-1.5B & Qwen2-1.5B & 0.526 & 0.465 & 0.555 & 0.627 & 0.706 & 0.641 & 0.751 & 0.803 \\
Qwen2-1.5B & GPT3.5 & 0.531 & 0.469 & 0.560 & \textbf{0.648} & 0.706 & 0.637 & 0.753 & 0.806 \\
Qwen2-7B & Qwen2-7B & \underline{0.539} & \underline{0.482} & \underline{0.566} & 0.634 & \textbf{0.724} & \textbf{0.672} & \underline{0.754} & 0.805 \\
Qwen2-7B & GPT3.5 & \textbf{0.543} & \textbf{0.487} & \textbf{0.570} & \underline{0.637} & \textbf{0.724} & \textbf{0.672} & \underline{0.754} & 0.807 \\
\bottomrule
\end{tabular}
\caption{Results for ablation studies with different combinations of LLMs in the Reasoning and Re-ranking stages. We adopt the best base KGC model, namely NBF-Net and SimKGC for the FB15k237 and the WN18RR datasets. }
\label{tab:ablation_llm}
\vspace{-0.2cm}
\end{table*}

\subsection{Main Results}
Table~\ref{tab:results} summarizes the performance of the $\text{KGR}^3$ framework on six different base KGC methods. The experiment results show that the best variant of $\text{KGR}^3$ significantly outperforms all baseline methods among all evaluation metrics. Compared to the previous state-of-the-art baseline, $\text{KGR}^3$, with Llama3-8B as the backbone LLM, achieves absolute Hits@$1$ improvements of $11.1\%$ and $4.0\%$ on the FB15k237 and the WN18RR datasets, respectively.~\footnote{$\text{KGR}^3$ obtains better performance with Qwen2-7B LLM. } It demonstrates that $\text{KGR}^3$ is highly effective for the KGC task. 

Notably, the improvement in Hits@1 is more substantial than that in Hits@3 and Hits@10. This indicates that the $\text{KGR}^3$ framework is particularly effective at identifying the most accurate answers. Since our framework primarily focuses on re-ordering top-$n$ entities from the initial ranked entity list, the upper bound of Hits@$1$, Hits@$3$, and Hits@$10$ are implicitly constrained by the Hits@$n$ performance of the base KGC model. Given that Hits@1 is typically further from this upper bound, the potential for improvement will be greater. Additionally, by leveraging semantic knowledge from entity contexts, the LLM gains a more comprehensive understanding of the entities, thereby enabling more precise inferences, particularly for top-ranked candidate answers. 

Compared to the selected base KGC models, the corresponding variants of $\text{KGR}^3$ consistently boost the performance on both datasets to a large margin. It shows that $\text{KGR}^3$ is compatible with various types of KGC models, confirming its strong generality and plug-and-play capability. In contrast, KICGPT fails to outperform the state-of-the-art GNN-based or text-based methods, which underutilizes the power of the LLM. In addition, variants of $\text{KGR}^3$ consistently outperform counterparts of the LLM-based baseline DIFT~\cite{DIFT} with the same base KGC models. The performance improvements can be primarily attributed to the incorporation of the reasoning module and the inclusion of entity descriptions for candidate answers. It should also be noted that DIFT necessitates an expensive re-training process and pre-trained KG embeddings to adapt to various KGC models~\cite{DIFT}. Without such a process, DIFT cannot guarantee the experimental results as claimed.

\subsection{Ablation Studies}
We verify the effectiveness of each component in the $\text{KGR}^3$ framework by answering the following research questions~(RQs). Table~\ref{tab:ablation_module} and ~\ref{tab:ablation_llm} shows the experimental results for ablation studies. 

\paragraph{RQ1: Does the reasoning module improve the final inference performance? }
To address this question, we simply apply the re-ranking module to candidate answers retrieved from the base KGC models. The performance drop in the ``$\text{KGR}^3$ w/o Reasoning'' variant demonstrates that the pre-trained LLM can provide plausible answers that are not initially ranked at the top positions by the base models. This capability effectively breaks through the limitations of base KGC models and increases the performance ceiling of our method.

\paragraph{RQ2: Whether entity descriptions contribute to enhancing KG completion?}
In the ``$\text{KGR}^3$ w/o Entity Descriptions'' variant, we remove the descriptions for the known entity of a query triple and each of its candidate answers, resulting in suboptimal experiment results. The performance decline re-confirms our hypothesis that LLMs may lack a fundamental understanding of certain entities within the KG, showing the importance of retrieving and leveraging textual contexts. 

\paragraph{RQ3: Can LLMs generate desirable KGC results without the help of KG triples? } 
We observe a significant performance decline with the ``$\text{KGR}^3$ w/o Neighbor Facts'' variant, particularly for Hits@$1$. 
In this case, the LLM can only rely on limited semantic knowledge derived from entity descriptions and its inherent knowledge base, which proves insufficient for generating precise predictions. This underscores that KG triples provide accurate and irreplaceable structural knowledge that is not inherently present in the LLM. In general, experimental results in Table~\ref{tab:ablation_module} reconfirm the complementary relationship between textual and structural contexts. 

\paragraph{Case study. } In addition, we conduct a case study on the hallucination case that may happen during the reasoning stage.  For example, when being instructed to complete the incomplete triple (<missing-entity>, \textit{/olympics/olympic\_games/parti-cipating\_countries}, \textit{/m/04vjh} [Mauritania]) which queries the Olympic Games in which Mauritania has participated, the pre-trained LLM (GPT) incorrectly responds “\textit{Mauritania has never participated in the Olympic Games}”. This error highlights how hallucinations can arise from the model's lack of critical factual knowledge.

Nevertheless, triple (\textit{/m/04vjh} [Mauritania], \textit{/olympics/olympic\_participating\_country/athletes - /olympics/olympic\_athlete\_affiliation/Olympics}, /m/06sks6 [2012 Summer Olympics]) shows that Mauritania does attend 2012 Summer Olympic games, which can help the LLM to figure out Mauritania also participated in [2008 Summer Olympics] during the re-ranking stage. This case study also shows that access to accurate factual knowledge is essential for LLMs to perform reasoning tasks like KGC.

\paragraph{RQ4: Can $\text{KGR}^\mathbf{3}$ ensure desirable performance with different LLMs?}
Experimental results in Table~\ref{tab:ablation_llm} show that $\text{KGR}^3$ consistently outperforms all baseline methods across all $8$ LLM combinations. Notably, $\text{KGR}^3$ achieves the state-of-the-art performance when using Qwen2-7B and GPT3.5 as the LLMs of the re-ranking and the reasoning module. Compared to variants using pre-trained open-source LLMs, those employing GPT3.5 in reasoning stage produce better predictions, showcasing its stronger semantic understanding and instruction following capabilities. 

It should be noted that $\text{KGR}^\mathbf{3}$, when using the same LLM, Llama2-7B, still outperforms DIFT by a large margin. Hence, we cannot simply attribute the performance improvements to the power of LLMs. In addition, $\text{KGR}^3$ can also produce plausible predictions with an 1.5B model, which reduces the average SFT time from $28.3$h to $9.05$h, demonstrating its strong robustness and effectiveness in low-resource settings. 

\begin{table}
\centering
\resizebox{0.48\textwidth}{!}{
\begin{tabular}{lcccc}
\toprule
\textbf{Settings} & \textbf{MRR} & \textbf{Hits@1} & \textbf{Hits@3} & \textbf{Hits@10} \\
\midrule
Reordering (RotatE) & 0.382 & 0.293 & 0.417 & 0.559 \\
$\text{KGR}^\mathbf{3}$ (RotatE) & 0.456 & 0.400 & 0.476 & 0.569 \\
\midrule
Reordering (GIE) & 0.391 & 0.301 & 0.426 & 0.573 \\
$\text{KGR}^\mathbf{3}$ (GIE) & 0.463 & 0.400 & 0.485 & 0.581 \\
\bottomrule
\end{tabular}
}
\caption{Ablation Experiments on FB15k-237 dataset with different re-ranking strategies.}
\label{tab:RQ5}
\vspace{-0.25cm}
\end{table}

\begin{figure*}
    \centering
    \includegraphics[width=0.90\textwidth]{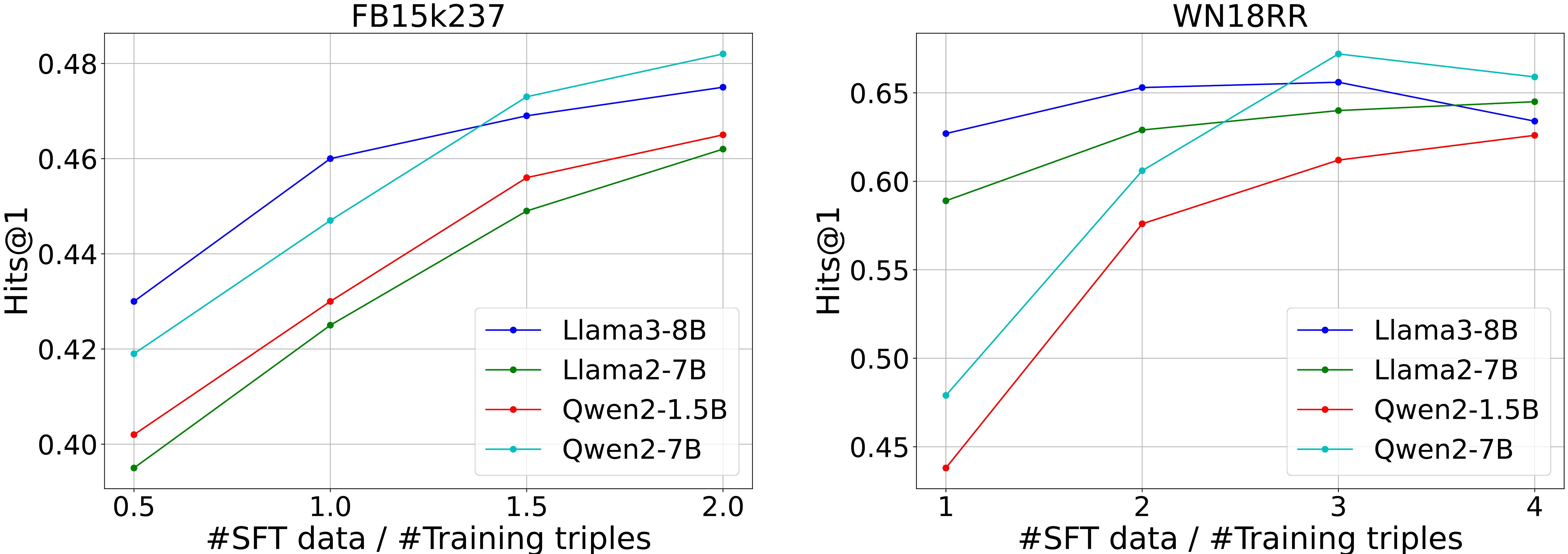}
    \caption{Hits@$1$ performance over the amount of SFT data on the two datasets with different LLMs.}
    \label{fig:epoch}
\end{figure*}

\paragraph{RQ5: How does the ranking strategy affect the performance of $\text{KGR}^\mathbf{3}$?} Rather than instructing the LLM to ``sort'' or ``reorder'' the entire candidate entity list, $\text{KGR}^\mathbf{3}$ guides the LLM to identify and place the most likely candidate at the top, while preserving the relative order of the remaining candidates. As shown in Table~\ref{tab:RQ5}, we conduct additional experiments to compare these two ranking strategies. Under the reorder setting, the LLM is fine-tuned to follow the original ranking produced by the corresponding base KGC models. However, the results remain suboptimal, suggesting that LLMs are not well-suited for sorting entire entity lists. This limitation can be attributed to the nature of KGs, which do not inherently provide ground truth rankings of entities for each query triple. As a result, it is not intuitive to claim that ground truth tail entity A is more relevant than ground truth tail entity B for a given query $(h, r, ?)$. Additionally, the experimental results demonstrate that constraining LLMs to mimic the ranking patterns of prior KGC methods does not alleviate their inherent limitations.

Furthermore, in some cases, LLMs may select an incorrect, or a false negative candidate answer, which reduces the rankings of the ground truth entity.
For example, during the re-ranking stage, the Llama3-8B model suggests “\textit{Solihull}” as the answer of query triple (\textit{England}, \textit{location contains}, \textit{<missing-entity>}), where the ground truth is “\textit{Pontefract}”. In fact, both “\textit{Solihull}” and “\textit{Pontefract}” are towns in England. Due to the inherent defect of the KG, the former is judged as a wrong answer. The re-ranking strategy adopted in $\text{KGR}^3$ ensures that when an LLM makes an incorrect judgment, the ranking of the ground truth answer drops by no more than 1 position, and hence, results in desirable performance improvements.

\paragraph{RQ6: How does the volume of SFT data affect the performance of $\text{KGR}^\mathbf{3}$?} 
From Figure~\ref{fig:epoch}, we can conclude that increasing SFT data generally improves the model performance. We attribute the boundary effects observed in the WN18RR dataset to the characteristics of its entities, which consist of common English words that can be readily interpreted by LLMs. It is also noteworthy that, even with a limited number of training samples, $\text{KGR}^{3}$ still achieves desirable results. 
Given the substantial performance improvements, taking additional computational costs is deemed justifiable. 

\section{Conclusion}
In this paper, we propose $\text{KGR}^3$, an LLM-based context-enriched KGC framework with three modules: Retrieval, Reasoning, and Re-ranking. By leveraging contextual information, $\text{KGR}^3$ effectively bridges the semantic gap between structural KG triples and natural language. Experimental results show that incorporating supporting triples and entity descriptions with LLM in-context learning and SFT significantly improves the KGC performance.  Future work will focus on adapting LLMs to other KG reasoning tasks such as inductive KGC and knowledge-based question answering. 

\section*{Acknowledgement}
The work described in this paper was partially supported by the Research Grants Council of the Hong Kong Special Administrative Region, China (CUHK 14222922, RGC GRF 2151185). We would also like to express our sincere gratitude to the reviewers and area chairs for their constructive comments and suggestions. 

\section*{Limitations}
Although the proposed framework achieves a significant breakthrough in the KGC task, it still has some remaining issues to be resolved in the future. Firstly, the proposed $\text{KGR}^3$ framework is not capable of handling the KGC task under an ``inductive setting''. The KGC task discussed in this paper, along with most related works, operates under a ``transductive setting'', where entities in test triples also exist in the training set. We plan to tackle unseen entities that are not present in the KG in the future. Secondly, the commonly adopted evaluation metrics Hits@$k$ necessitate ranking the plausibility of all entities within the KG for each query triple. Due to the limitations imposed by the maximum sequence length of LLMs, it is impractical to rank tens of thousands of entities within a KG. To \textit{meet the evaluation criteria}, one may have to select a proper subset of entity candidates using existing KGC models. Hence, the base KGC model may inevitably constrain the final performance of the proposed framework. 

\section*{Ethics Statement}
To the best of our knowledge, this work does not involve any discrimination, social bias, or private data. All the datasets are constructed from open-source KGs such as Wikidata.
Therefore, we believe that our study complies with the ACL Ethics Policy.

\bibliography{custom}

\clearpage

\appendix
\onecolumn
\section{Appendix}
\label{prompts}

\subsection{Prompt templates of Reasoning stage}\label{sec:appendix2}
Table~\ref{tab:pt_reasoning} shows the prompt templates of the \textbf{Reasoning} stage and provides an example.
\begin{table}[h!]
    \centering
    \resizebox{\textwidth}{!}{
    \begin{tabular}{p{\textwidth}} 
        \toprule
        \textcolor{darkblue}{\textbf{\#\# In-context Demonstrations:}} \\
        \textbf{User}: \textcolor{red}{Washington County: county in Pennsylvania, U.S.} The question is to predict the head entity [MASK] from the given ([MASK], \textit{location adjoining\_relationship adjoins}, Washington County) by completing the sentence 'Washington County is the adjoins of what location? The answer is '.\\
        \textbf{Assistant}: The answer is Westmoreland County, so the [MASK] is Westmoreland County. \textcolor{red}{Westmoreland County: county in Pennsylvania, United States} \\
        \\
        \textbf{User}: \textcolor{red}{Rockland County: suburban county in New York, United States} The question is to predict the head entity [MASK] from the given ([MASK], \textit{location adjoining\_relationship adjoins}, Rockland County) by completing the sentence 'Rockland County is the adjoins of what location? The answer is '.\\
        \textbf{Assistant}: The answer is Bergen County, so the [MASK] is Bergen County. \textcolor{red}{Bergen County: county in New Jersey, United States} \\
        \midrule
        \textcolor{darkblue}{\textbf{\#\# Query triple}:}
        ([MASK], \textit{/location/adjoining\_relationship/adjoins}, Champaign) \\
        \midrule
        \#\# \textbf{\textcolor{darkblue}{Reasoning prompts}:} \\
        \textbf{User}: The question is to predict the head entity [MASK] from the given ([MASK], \textit{location adjoining\_relationship adjoins}, Champaign) by completing the sentence 'Champaign is the adjoins of what location? The answer is '. Output all some possible answers based on your own knowledge, using the format '[answer1, answer2, ..., answerN]' and please start your response with 'The possible answers:'. Do not output anything except the possible answers. \\
        \\
        
        \#\# \textbf{\textcolor{red}{Context-aware} \textcolor{darkblue}{Reasoning}:} \\
        \textbf{User}: Here are some materials for you to refer to: \\
        \textcolor{red}{Champaign: Champaign is a city in Champaign County, Illinois, United States. The population was 88,302 at the 2020 census. It is the tenth-most populous municipality in Illinois and the fourth most populous city in the state outside the Chicago metropolitan area. It is a principal city of the Champaign–Urbana metropolitan area, which had 236,000 residents in 2020. } \\
        \\
        The question is to predict the head entity [MASK] from the given ([MASK], \textit{location adjoining\_relationship adjoins}, Champaign) by completing the sentence 'Champaign is the adjoins of what location? The answer is '. Output all the possible answers you can find in the materials using the format '[answer1, answer2, ..., answerN]' and please start your response with 'The possible answers:'. Do not output anything except the possible answers. If you cannot find any answer, please output some possible answers based on your own knowledge. \\
        \midrule
        \#\# \textcolor{darkblue}{\textbf{Reasoning results}:} \\
        \\
        \textbf{Assistant: }The possible answers: [Urbana, Champaign County, Illinois Silicon Prairie, Parkland College]\\ 
        \bottomrule
    \end{tabular}
    }
    \caption{Prompt Template of context-aware reasoning.}
    \label{tab:pt_reasoning}
\end{table}

\clearpage
\subsection{Prompt templates of Re-Ranking stage}\label{sec:appendix3}
Table~\ref{tab:pt_ranking} shows the prompt templates of the \textbf{Re-ranking} stage and gives an example. It is noteworthy that this case empirically shows the effectiveness of the \textbf{Reasoning} and \textbf{Re-ranking} processes. The ground truth answer ``Urbana'' does not rank in a leading position by the KGC model. However, the LLM provides plausible candidates including the ground truth answer ``Urbana'', by analyzing the context of the known entity ``Champaign'' in the incomplete triple during the \textbf{Reasoning} process. In addition, the fine-tuned LLM succeeds in selecting the correct answer from the candidate list based on relevant textual contexts and neighbor facts during the \textbf{Re-ranking} process. 
\begin{table}[h!]
    \centering
    \resizebox{\textwidth}{!}{
    \begin{tabular}{p{\textwidth}} 
        \toprule
        \textcolor{darkblue}{\textbf{\#\# Query triple}:}
        (< ??? >, \textit{/location/adjoining\_relationship/adjoins}, Champaign) \\
        \midrule
        \#\# \textbf{\textcolor{darkblue}{Re-Ranking pompts}:} \\
        \textbf{User}: Here is an incomplete triple with missing head entity <missing-entity>: (<missing-entity>, location location adjoin\_s. location adjoining\_relationship adjoins, Champaign). \\
        \\
        Following are \underline{some contexts about tail entity Champaign}: \\
        \textcolor{red}{Champaign is a city in Champaign County, Illinois, United States.} The population was 88,302 at the 2020 census. It is the tenth-most populous municipality in Illinois and the fourth most populous city in the state outside the Chicago metropolitan area. It is a principal city of the Champaign-Urbana metropolitan area, which had 236,000 residents in 2020. \\
        Following are \underline{some triple facts of entity Champaign}: \\
        (Illinois, location location contains, Champaign) \\ 
        (Ludacris, people person place of birth, Champaign)\\
        (Champaign County, location location contains, Champaign) \\
        \\
        Please select the most appropriate entity for <missing-entity> from the candidate answer list: \\
        Champaign County: county in Illinois, United States, \\
        McHenry County: county in Illinois, United States, \\
        Lake County: county in Illinois, United States, \\
        Cook County: county in Illinois, United States, \\
        Madison County: county in Illinois, United States, \\
        St. Clair County: county in Illinois, United States, \\
        DuPage County: county in Illinois, United States, \\
        McLean County: county in Illinois, United States, \\
        Champaign: city in Champaign County, Illinois, United States, \\
        Peoria County: county in Illinois, United States of America, \\
        Kane County: county in Illinois, United States, \\
        Tazewell County: county in Illinois, United States, \\
        Oak Park: village in Cook County, Illinois, United States; suburb of Chicago, Illinois, \\
        Will County: county in Illinois, United States, \\
        Lake Forest: city in Lake County, Illinois, United States \\
        Springfield: city in and county seat of Sangamon County and Illinois federated state capital city \\
        Aurora: city in Kane County, Illinois, United States \\
        \textbf{Urbana}: \underline{town in and county seat of Champaign County, Illinois, United States} \\
        Alton: city in Madison County, Illinois, United States \\
        Kankakee: city in Kankakee County, Illinois, United States. \\
        \midrule
        \#\# \textcolor{darkblue}{\textbf{Re-Ranking Results}:} \\
        \textbf{Assistant}: Urbana \\
        \#\# \textcolor{darkblue}{\textbf{Evaluation}:} The ground truth ``Urbana'' hits at 1.\\
        \bottomrule
    \end{tabular}
    }
    \caption{Prompt Template of context-aware ranking.}
    \label{tab:pt_ranking}
\end{table}

\section{Implementation Details}~\label{sec:implementation} 
We conduct all our experiments on a Linux server equipped with $2$ Intel Xeon Platinum 8358 processors and $8$ A100-SXM4-40GB GPUs.~\footnote{Only 2 GPUs are used in our experiments. } For TransE and CoLE, we adopt the candidate entities provided in~\cite{DIFT}. For other base KGC methods, candidate entities are retrieved using their publicly available implementations. We tested our proposed method using $4$ different LLMs, specifically Llama2-7B, Llama3-8B, Qwen2-1.5B, and Qwen2-7B. The time costs of SFT and re-ranking are reported in Table~\ref{timecomplexity}. It is noteworthy that the duration of the experiments is subject to specific configurations, which can be further reduced through parallel inference or the application of vLLM service. Additionally, we also evaluated OpenAI's gpt-3.5-turbo-0125 model in the reasoning stage.
During the re-ranking stage, we employ the LLaMA-Factory framework~\cite{llamafactory} to fine-tune the mentioned LLMs. For LoRA adjustments, we set the rank to 16 and the alpha value to 32. We utilize the AdamW optimizer~\cite{adamW} to train our model, with an initial learning rate of $1.0e\text{-}4$, a per-device batch size of $2$ and a gradient accumulation step of $4$ iterations. We adopted BF$16$ precision to reduce the GPU memory usage. For each query triple, we retrieve $3$ supporting triples for in-context demonstration and provide $n=20$ candidate answers (during training, we provide $1$ ground truth entity and $19$ negative samples) to the LLM. Detailed hyperparameter settings are listed in Table~\ref{hyperparameters}. 

\begin{table}[htbp]\centering
\small
\begin{tabular}{cc}
\toprule
\textbf{Hyperparameters} & \textbf{Settings} \\ 
\midrule
SFT learning rate & 1e-4 \\ 
Per-device batch size & 2 \\ 
\# GPUs used & 2 \\ 
Gradient accumulation step & 4 \\ 
LoRA rank & $16$ \\
LoRA $\alpha$ value & $32$ \\
\hline
Re-ranking scope $n$ & $20$ \\ 
\# Easy negative samples & $9$ \\ 
\# Hard negative samples & $10$ \\ 
$\delta$ & $50$ \\ 
$p$ & $10$ \\
\bottomrule
\end{tabular}
\caption{Hyperparameter settings.}
\label{hyperparameters}
\end{table}

\begin{table}[htbp]\centering
\small
\begin{tabular}{lccccc}
\toprule
\textbf{Dataset} & \textbf{Qwen2-1.5B} & \textbf{Qwen2-7B} & \textbf{Llama2-7B} & \textbf{Llama3-8B} & \textbf{Test (Llama3-8B)} \\
\midrule
FB15k237 & 12h 12min & 36h 50min & 43h 42min & 38h 27min & 0.180s / query \\
WN18RR & 5h 54min & 22h 33min & 19h 30min & 18h 12min & 0.152s / query
\\
\bottomrule
\end{tabular}
\caption{Time costs of supervised fine-tuning and re-ranking. }
\label{timecomplexity}
\end{table}


\end{document}